\documentclass[runningheads]{llncs}
\usepackage[T1]{fontenc}
\usepackage{cleveref}
\usepackage{tikz}
\usepackage{graphicx}
\usepackage{xspace}
\usepackage{tabularx}
\newcommand{\etal}{\emph{et al.}\xspace}
\begin{document}
\title{Future Video Prediction from a Single Frame for Video Anomaly Detection}
\author{Mohammad Baradaran\inst{1}\orcidID{0000-0003-0246-4370} \and
\\
Robert Bergevin\inst{2}\orcidID{0000-0002-1115-7471} }
%
%\authorrunning{M. Baradaran et al.}
\institute{Université Laval, 065, avenue de la Medecine Quebec Canada, G1V 0A6
\email{mohammad.baradaran.1@ulaval.ca}\\
\and
Université Laval, 065, avenue de la Medecine Quebec Canada, G1V 0A6\\
\email{robert.bergevin@gel.ulaval.ca}}
\maketitle              % typeset the header of the contribution
\begin{abstract}
Video anomaly detection (VAD) is an important but challenging task in computer vision. The main challenge rises due to the rarity of training samples to model all anomaly cases. Hence, semi-supervised anomaly detection methods have gotten more attention, since they focus on modeling normals and they detect anomalies by measuring the deviations from normal patterns. Despite impressive advances of these methods in modeling normal motion and appearance, long-term motion modeling has not been effectively explored so far. Inspired by the abilities of the future frame prediction proxy-task, we introduce the task of future video prediction from a single frame, as a novel proxy-task for video anomaly detection. This proxy-task alleviates the challenges of previous methods in learning longer motion patterns. Moreover, we replace the initial and future raw frames with their corresponding semantic segmentation map, which not only makes the method aware of object class but also makes the prediction task less complex for the model. Extensive experiments on the benchmark datasets (ShanghaiTech, UCSD-Ped1, and UCSD-Ped2) show the effectiveness of the method and the superiority of its performance compared to SOTA prediction-based VAD methods.

\keywords{Video anomaly detection\and long-term motion modeling\and future video prediction\and semi-supervised learning.}

\end{abstract}

\section{Introduction}
Videos are rich sources of information and can be used to extract valuable insights about the world around us. With the widespread availability of cost-effective and high-quality cameras, there has been a surge in the amount of video data being recorded. Video analysis is of great importance for various applications (e.g., security, traffic monitoring, etc.).One crucial piece of information that can be obtained from video streams is the possible presence of abnormal events, as their occurrence often necessitates swift action. Abnormal video events refer to events that deviate significantly from expected patterns (i.e. normals), in a given context.

Many researchers in the field of video anomaly detection have put forth deep-learning-based methods, drawing inspiration from the success of Deep Neural Networks (DNN) in numerous computer vision applications such as segmentation and classification. Researchers have endeavored to effectively model and analyze both spatial and temporal features to detect related anomalies. Given the challenge of acquiring enough training samples for anomalies, they have shown a general interest in semi-supervised VAD methods, in which the objective is to fit a normality model on normal events and to detect anomalies by measuring deviations from the learned model. These methods usually leverage unsupervised DNNs aiming to learn a model for a proxy-task on normal data. A proxy task is defined as a task that is not directly linked to the anomaly detection task but can be utilized for anomaly detection by measuring the performance of the model on anomalies if it has been trained solely on normal data. Frame reconstruction and future frame prediction are the most frequently used proxy-tasks for VAD.

While existing state-of-the-art VAD methods have achieved impressive results in modeling normals and detecting related anomalies, they suffer from a common shortcoming: unlike the local short-term motion patterns, long-term motion pattern has not been effectively explored and considered.  In essence, exisitng methods take into account instant motion features (e.g., instant speed, instant direction, or motion changes) for anomaly detection. In this paper, we propose a model that can effectively incorporate long-term motion information for video anomaly detection, in addition to short-term motion patterns. The proposed method takes a prediction-based approach, inspired by earlier methods \cite{b14,b31,b35} that predict the future, based on past data. However, unlike the existing methods, our method predicts future video instead of only the immediate future frame. Additionally, unlike methods such as \cite{b41,b43}, and drawing inspiration from \cite{paper1}, the future video (the next 10 frames) is predicted from an initial frame rather than a sequence of past frames.

Since predicting future frames solely based on a single initial frame may confuse the model in estimating the correct motion direction and lead to incorrect predictions, the proposed method includes the initial motion direction as an additional input, along with the initial frame. Furthermore, prior research \cite{paper2,paper3} has demonstrated that deterministic video prediction results in blurry outcomes, which are essentially the mean of all possible outcomes. Therefore, in the proposed method, a separate branch that utilizes a Variational Autoencoder (VAE) is employed, to make the generation task stochastic. In the proposed method, two distinct encoders are utilized for encoding the necessary inputs into a latent space, and a shared decoder is employed to decode them into future frames. In summary, the proposed method introduces the following contributions:

- A video anomaly detection approach that integrates long-term motions for improved anomaly detection.\\
- The first method to predict semantic future video from an initial semantic frame for video anomaly detection.\\
- Superior performance in comparison to SOTA prediction-based methods.%, as supported by quantitative results.

\section{Related work}

Semi-supervised VAD methods have received significant attention from researchers in the field, as abnormal samples are not always readily available, unlike normal samples. In semi-supervised video anomaly detection methods, one or multiple proxy-tasks are used to train an unsupervised model on normal samples for a given task. The anomaly score is then calculated by measuring the deviation of normal models on each test sample.

One commonly used proxy-task in semi-supervised VAD methods is frame reconstruction \cite{b23,b29,b1,b22,b47} in which, an unsupervised neural network is trained on normal frames, assuming that the reconstruction error would be comparatively higher for abnormal frames. The main limitation of these methods is the difficulty to effectively incorporate motion information into the anomaly detection decision, mainly due to the dominance of appearance features on temporal ones \cite{memo5}. Even with the incorporation of motion-aware layers (such as Conv-LSTM-AE) \cite{b12,b36}, motion information is not effectively modeled and considered in these methods. To this end, researchers have proposed two/multi-stream \cite{b25,b19} methods, where spatial and temporal features are modeled separately, to have an independent effect on the final decision, rather than being dominated by the other one. Anomaly detection in the motion branch is generally accomplished by extracting explicit motion features from frames (such as optical flow maps or temporal gradients of frames) and leveraging CNNs' ability in analyzing spatial data to analyze motion patterns \cite{b17}.

Neygun et al \cite{b46a} utilized a generative adversarial network to map a raw frame to its corresponding optical flow map for motion information modeling. However, predicting motion direction from a single frame is challenging for the network, leading to incorrect predictions. To address the mentioned challenge, Baradaran et al \cite{memo1} proposed translating from the raw frame to its corresponding optical flow magnitude map for motion anomaly detection. They also proposed an additional method \cite{memo2} that utilizes multiple attention networks to incorporate contextual information for more accurate motion modeling.

\begin{figure}[hbt!]
\centering
\includegraphics[width=10cm]{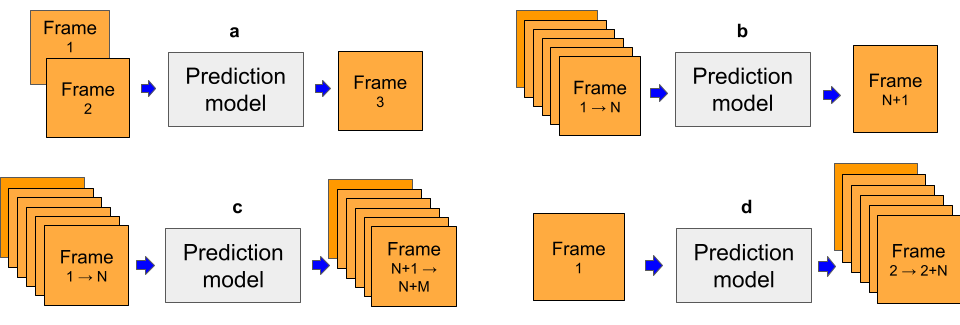}
\caption{Different prediction-based VAD methods. (a) Video-to-Frame (2:1). (b) Video-to-Frame (N:1). (c) Video-to-Video (N:N). (d) Frame-to-Video (1:N).} \label{fig1}
\end{figure}

As a proxy-task customized for motion modeling, researchers have proposed future frame prediction \cite{b31,b35,b11,b38,b4j,b27j,b42j,b47j,b61,b57,b52}. Prediction-based models are trained to predict future frames by observing a history of past frames and are expected to encounter difficulties in predicting the location of objects with abnormal motion, in future frames. Baradaran et al \cite{memo1} predict the immediate next frame to detect abnormal motions. Other researchers \cite{b14,b31,b35,b42j} propose to feed a longer sequence as an input (generally 4 frames) and design sophisticated networks to effectively predict the immediate next frame. However, both groups (i.e., 2:1 and N:1 as in Fig.1 a,b) have shortcomings; mainly long-term motion is not effectively modeled. Even by increasing the length of historical frames, models tend to consider the last frames for future prediction, due to the high correlation between adjacent frames. To consider longer motion patterns for VAD, \cite{b41,b43} have proposed Video-to-Video prediction proxy-task (N:N as in Fig.1 c). Despite showing better results in considering longer motion patterns, they fail in detecting monotonic abnormal motions (such as a car driving in a constant fast motion through frames), since the network can infer the motion pattern from historical frames to predict the future frames, which may result in a precise prediction for anomalies. These methods are not fully explored apparently due to the complexity of the task.

Motivated by the ability of future video prediction task in modeling longer motion patterns, we propose a novel proxy-task for video anomaly detection and introduce the Frame-to-Video prediction task (Fig1. d) to video anomaly detection. Moreover, unlike conventional Frame-to-Video applications, we provide  the semantic segmentation map of initial frames to predict the semantic map of future frames. Replacing raw frames with corresponding semantic maps not only makes the method aware of the object classes but also addresses the challenges of the model in dealing with the complexity of the background in raw frames.
\vspace{-2mm}

\section{Method}
\vspace{-1mm}
\subsection{Overview of the proposed method}

We propose a novel video anomaly detection method that addresses the challenges of existing methods by incorporating long-term motion patterns, in addition to short-term motions. Our approach is based on predicting future video from a single initial frame, leveraging the semantic map of each frame instead of the raw frame as input and target to reduce prediction task complexity. This is the first work to introduce future video prediction from an initial frame for video anomaly detection. The overview of the proposed method is depicted in Fig.2.

\begin{figure}
\centering
\includegraphics[width=10cm]{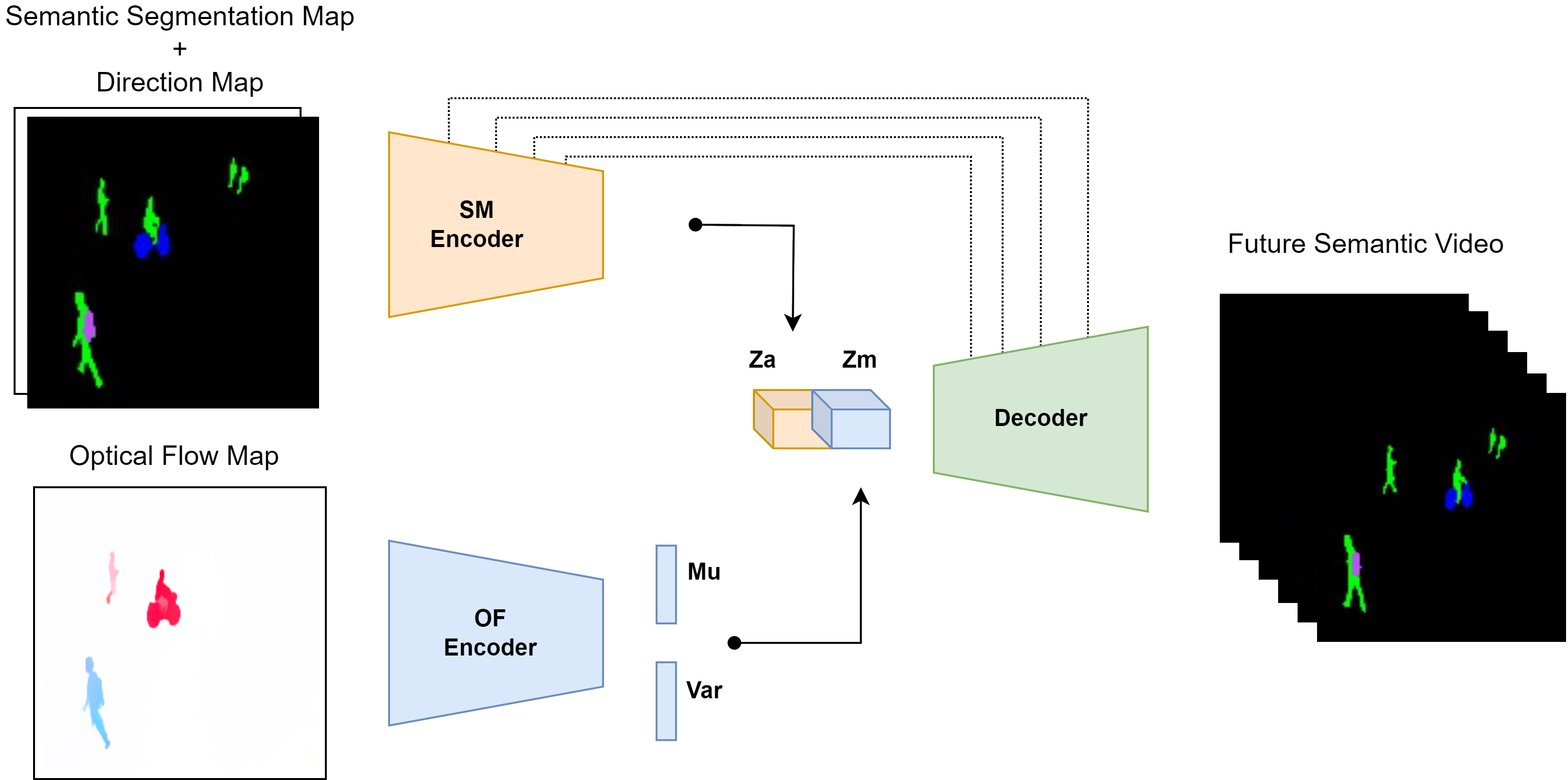}
\caption{Overview of the proposed Frame-to-Video prediction based VAD method.} \label{fig2}
\vspace{-3mm}
\end{figure}

\subsection{Video-to-Video}

Previous studies have shown that VAD methods based on predicting the next frame from a sequence of previous frames face a few challenges: 1) They may generalize to anomalies due to the high learning capacity of DNNs. 2) Since these methods tend to use the last frame(s) for inference, only short-term motions are considered for VAD. To overcome this challenge, in the first phase of our experiments, we designed a video-to-video prediction-based VAD method. To achieve this goal we leveraged the SimVP (Simpler yet Better Video Prediction) model \cite{simvp}, one of the state-of-the-art video prediction models, that  receives a sequence of past raw frames and predicts future raw frames.

\subsection{Frame-to-Video}
Experiments on Video-to-Video methods have shown that they suffer from a shortcoming: DNNs can generalize to anomalies, as they can infer the trend of anomalous motions from past frames. In other words, DNNs can formulate the prediction as finding the relation between frames rather than learning patterns. To address this, we formulated our VAD method as Frame-to-Video prediction. Our generative model consists of one convolutional encoder, one convolutional variational encoder, and a shared decoder, and takes an initial frame as input to generate the future video conditioned on initial motion information.

\subsection{Stochasticity}
To address the challenge of generating sharp video frames, which is a common challenge in deterministic prediction models, we use a variational encoder in our lower branch \cite{paper3}. This is in line with previous works \cite{paper2}, which have shown that deterministic models can underperform in this task. By adopting a variational encoder, we can model the inherent stochasticity of the problem, capturing multiple possible future motions that are conditioned on the motion information in the initial frame.

\subsection{Semantic maps}
Our initial experiments indicated that utilizing raw RGB frames as input and target may contribute to the complexity of the video prediction task. The abundance of spatial details in raw frames requires the predictor to focus on learning and reconstructing these details rather than learning temporal patterns. This becomes more evident when considering backgrounds which usually occupy most of the frame and have no motion, yet consist of many details. To address this, we conducted experiments using semantic maps as input and target. The results showed that utilizing semantic maps makes it easier for the model to converge and learn motion patterns. This also helps address previous challenges, such as the complexity of backgrounds, as semantic colors replace the previous complex information (textures, intensities, shapes, etc.).

Previous studies, such as \cite{paper1}, emphasize the importance of leveraging semantic information for predicting motion. These works argue that different object classes have very distinctive motions, and utilizing semantic information can aid in better modeling these motions. Results from these works demonstrate that with semantic representation as input, the model can learn better motion for dynamic objects compared to not utilizing semantic information. 

\subsection{Initial motion direction}
Our experiments also showed that predicting video from a static frame resulted in wrong motion directions, as the model favored the most frequent directions and preferred to predict motions mostly in these directions. To address this issue, we incorporated the initial motion direction along with the initial frame in the prediction process. We introduced the direction information to the model at the lowest layer, by concatenating the frame and the direction map to form the input for the model.  This provides exact direction information for each pixel in the frame. The direction of motion was provided through two separate maps representing the Sine and Cosine of the motion angle, which was extracted using a pre-trained optical flow extraction model called Raft \cite{raft}.

\begin{equation}
Mag, Ang = OF(I_{t-1}, I_{t})
\end{equation}

\begin{equation}
DM = Concatenate( |Cos(Ang)| , |Sin(Ang)|)
\end{equation}
Where, \verb'DM', \verb'Mag', and \verb'Ang' stand for the direction map, magnitude, and angle of motion respectively.
\subsection{Network}
Our proposed architecture consists of two encoders and a shared decoder, all utilizing 3x3 2D convolutional filters. The first encoder, referred to as the Semantic Map (SM) Encoder in Fig.2, is a convolutional encoder that takes the concatenation of the initial semantic segmentation map and the initial direction map as input. It encodes these inputs into 512x4x4 feature maps. The second encoder, known as the Optical Flow (OF) Encoder, is a variational convolutional encoder that encodes the initial optical flow map into a normal distribution. From this distribution, a motion feature map of size 512x4x4 is sampled. The encoders downsize the feature resolution using 2x2 max pooling.

On the other hand, the shared decoder is a convolutional decoder that receives the concatenated encoded features and decodes them, using transpose convolution layers, to generate future semantic segmentation maps. Skip connections are incorporated between the convolutional encoder and the decoder to facilitate the transfer of information from the encoder to the decoder while preserving the original high-resolution features. Throughout the architecture, Leaky ReLU with a slope of 0.1 is utilized as the activation function for all layers, except for the last layer of the decoder, which employs the sigmoid activation function.

\subsection{Loss functions}

To train our proposed Frame-to-Video-based VAD method, we leverage three loss functions. As a primary loss function used in prediction-based methods, we use reconstruction loss, denoted as $L_{Rec}$. Let $y\string^$ be the predicted future video and $y$ the expected future video, we minimize the $L2$ distance between $y$ and $y\string^$ as the reconstruction loss:

\begin{equation}
L_{Rec}= ||y-y\string^||_{2}^2
\end{equation}

We also define a temporal gradient loss, to ensure that the evolution between frames (i.e., motion patterns) has been considered for model optimization. This loss function is defined as below:

\begin{equation}
L_{TG} = MAELoss(TG(y), TG(y\string^))
\end{equation}

Where TG is a function that computes the temporal gradient of the video and is defined as below:

\begin{equation}
TG([F_{1},F_{2},..., F_{N}])=[F_{2}-F_{1},F_{3}-F_{2}, …, F_{N}-F_{N-1}]
\end{equation}

Where $F_{t}$ denotes the frames \verb't' at different times in the frame sequence. Finally, KL-divergence is also considered as a complementary loss to add regularization to the training while also considering the prior normal distribution for the input optical flow maps to  address the stochasticity in future prediction.

Overall, the total training loss for the model is defined below:

\begin{equation}
L_{Total}=L_{Rec}+L_{TG}+\beta*KL
\end{equation}

In this equation, a controlling hyperparameter named $\beta$ is added to weigh the KL-divergence in the final loss value.

\subsection{Inference}
During the inference phase, we use the reconstruction loss value to calculate the anomaly score, assuming that a model trained on normal video clips would generate higher reconstruction loss for anomalous videos compared to normal videos. By setting a threshold for the anomaly score, we can detect anomalous events in the test video. 

Considering the frame rate, we can assume that adjacent frames are similar to each other and there are no big changes from one frame to the immediate next frame. However, in our experiments, we observe some big fluctuations in anomaly scores between adjacent frames. Our analysis shows that these jumps can happen mostly due to failures of the semantic segmentation framework. Hence we apply the Savitzky–Golay filter \cite{savgol} as a temporal relaxing technique. Finally, calculated anomaly scores are normalized to the range of [0,1] using Eq.7.

\begin{equation}
L_{norm}= \frac{L_{Rec} - Min(L_{Rec})}{Max(L_{Rec}) - Min (L_{Rec})}
\vspace{-2mm}
\label{eq:eq3}
\end{equation}

\section{Experiments}

\subsection{Datasets}
The ShanghaiTech Campus \cite{b31}, UCSD-Ped1, and UCSD-Ped2 datasets\cite{ped} are extensively adopted reference in the evaluation of semi-supervised video anomaly detection methods. The training partitions of these datasets exclusively encompass normal frames, in contrast to the testing partitions, which incorporate both normal and abnormal instances. Across all these datasets, normality is characterized by individuals walking on the sidewalk, whereas any encounters with unobserved objects or motion patterns are defined as anomalies. The ShanghaiTech Campus dataset exhibits a higher degree of complexity relative to UCSD-Ped1 and Ped2, featuring 13 different scenes and considerably expanded types of anomalies. However, the UCSD datasets, notably Ped1, present their own distinct complexities owing to their lower resolution and grayscale frames. These features potentially lead to occasional segmentation inaccuracies.
\vspace{-4mm}
\subsection{Evaluation metric}

In order to assess the effectiveness of our proposed approach, we followed prevalent SOTA methods in the field and employed the frame-level Area Under Curve (AUC) metric for evaluation. The AUC curve is constructed by recording various True Positive Rates (TPR) and False Positive Rates (FPR) of the approach, brought about by changing the anomaly score threshold from its minimum to its maximum limit. A higher AUC value signifies a better performance level.
\vspace{-3mm}
\subsection{Implementation details}
Through all stages of our experiments, which encompassed both Video-to-Video and Frame-to-Video methods, we standardized the images to dimensions of 128x128 pixels. For the Video-to-Video based VAD method, which use SimVP as a generator, we followed the guidance provided in the main reference (\cite{simvp}) for training, also supplying ten frames as input and setting an equal number of frames as targets. During our Frame-to-Video experiment, we employed Mask-RCNN, pre-trained on MS COCO, for the generation of semantic maps for each individual frame. Following the resizing of these maps to 128x128 pixels, a single frame is designated as input and its subsequent ten frames are treated as targets for prediction. For the computation of optical flow maps, we utilized the RAFT optical flow extraction model. We select its basic version to facilitate a swifter processing time. All inputs and corresponding targets are supplied to the model in batches of 16. To facilitate the training of the model, learning rates are set at an initial value of 1-e3 and were halved at intervals of every 10 epochs. Parameter optimization is carried out using the Adam optimizer. Finally, the best results are obtained for $\beta=1$.

\subsection{Results of Video-to-Video experiments}
Experiments (Table1) show that formulating VAD as prediction of farther frames boosts the anomaly detection frame-level accuracy, as going far to the future the difference between predictions and expectations (real future frames) grows in anomalies compared to normals.

\begin{table}
\centering
\caption{Performance (AUC) of the Video-to-Video prediction-based VAD (with SimVP as predictor) on ShanghaiTech dataset for different prediction time steps in the future.}\label{tab1}
\begin{tabular}{|l|l|l|l|l|l|l|l|l|l|l|}
\hline
prediction time step &  1 & 2 & 3 & 4 & 5 & 6 & 7 & 8 & 9 & 10\\
\hline
AUC &  74.37 & 76.37 & 76.9 & 77.76 & 78.42 & 78.43 & 78.47 & 78.55 & 78.73 & 78.91
\\

\hline
\end{tabular}
\end{table}

However, related qualitative experiments (Fig.3) also depict that Video-to-Video based VAD methods may infer the anomalous motion patterns from past frames and use them to predict future frames, in which case even by going far to the future, the difference between predictions and expectations does not grow considerably for anomalies (anomalies are highlighted in red boxes in Fig.3).

\begin{figure}
\centering
\includegraphics[width=8cm]{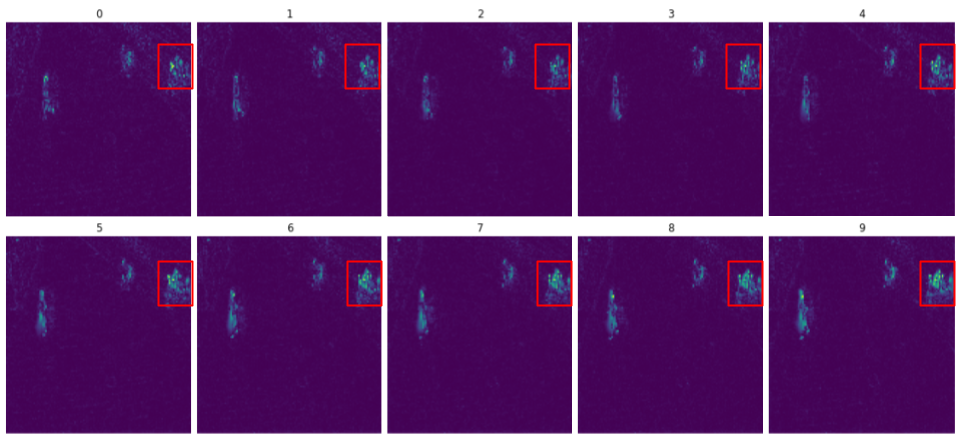}
\caption{Anomaly map (difference between predicted future frames and expected (real) future frames) of the Video-to-Video based VAD method for a sample clip from ShanghaiTech, for different time steps in the future.} \label{fig3}
\vspace{-3mm}
\end{figure}

\vspace{-3mm}

\subsection{Qualitative evaluation for Frame-to-Video method}
The qualitative results (Figure 4: anomaly maps of predictions in different time steps) provide compelling evidence that the proposed method is highly proficient in detecting unfamiliar (novel) objects accurately, primarily from the immediately following frame (Fig.4, first row). Besides, the method exhibits the ability to identify abnormal motions from the subsequent frames. Overall, the effectiveness of anomaly detection improves as we move further into the future. Row 2 and 3 in Fig.4 show some samples, which contain anomalous motion patterns (fighting and chasing here) by normal objects (i.e. humans in green). In this case, the anomaly map activations in the immediate next frame are not distinct for anomalies compared to normals. However, as we go  toward the future, the disparity between predictions and expectations grows considerably for anomalies compared to normals. Conversely, when examining normal objects with normal motion (Fig.4, last row), the anomaly maps display minimal activations through all frames, primarily confined to the boundaries of the objects. %It is worth noting that this issue of small activations at the object boundaries is a common challenge encountered in image reconstruction and segmentation tasks.
\vspace{-3mm}
\begin{figure}[hbt!]
\centering
\includegraphics[width=8cm]{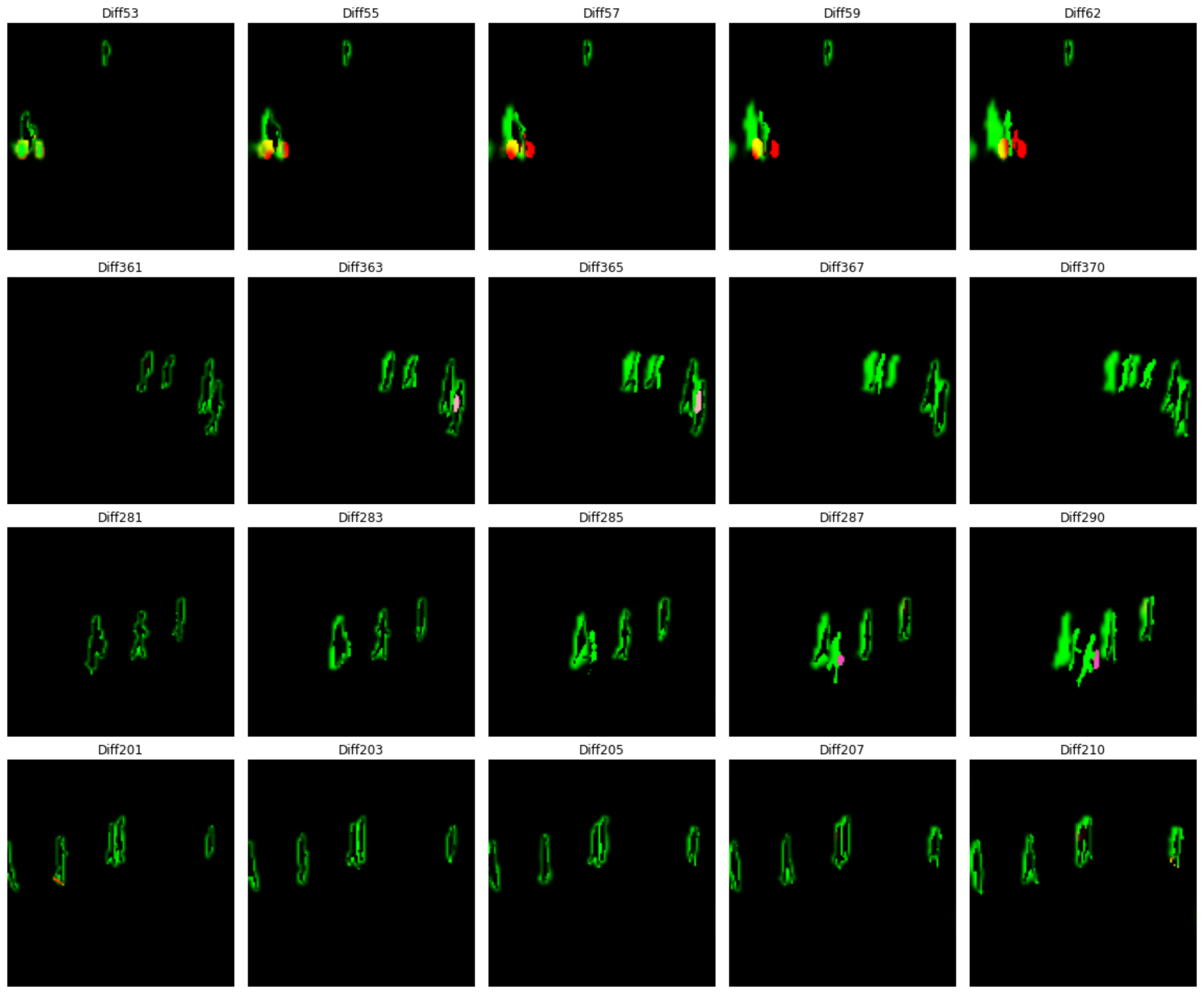}
\caption{Anomaly maps of different clips from ShanghaiTech dataset. Each row shows results for different clips and each column shows the anomaly maps for different frames in the future (columns 1,2,3,4,5 respectively are for the next 1st, 3rd, 5th, 7th, and 10th frames in the future). Different colors refer to different object classes and higher intensities in the anomaly maps indicate higher confidence in anomaly detection.} \label{fig4}
\vspace{-2mm}
\end{figure}

As another example, Fig.5 exhibits the initial frame in the first column and the future predictions in the next columns. In this figure, the first row shows the predictions of the model, and the second and third rows respectively illustrate the corresponding expectations (semantic segmentation of future frames as pseudo-ground-truth) and computed anomaly maps (difference between predictions and expectations) for each time sample. These qualitative results indicate that in the case of abnormal patterns, the disparity between predictions and expectations increases more rapidly when progressing toward future frames compared to normal patterns. Results show that novel objects (a bike in this example: segmented in red in the pseudo-ground-truth) are classified as human (green color) in the predictions, as humans have been frequently observed during the training.
\vspace{-1mm}
\begin{figure}[hbt!]
\centering
\includegraphics[width=12cm]{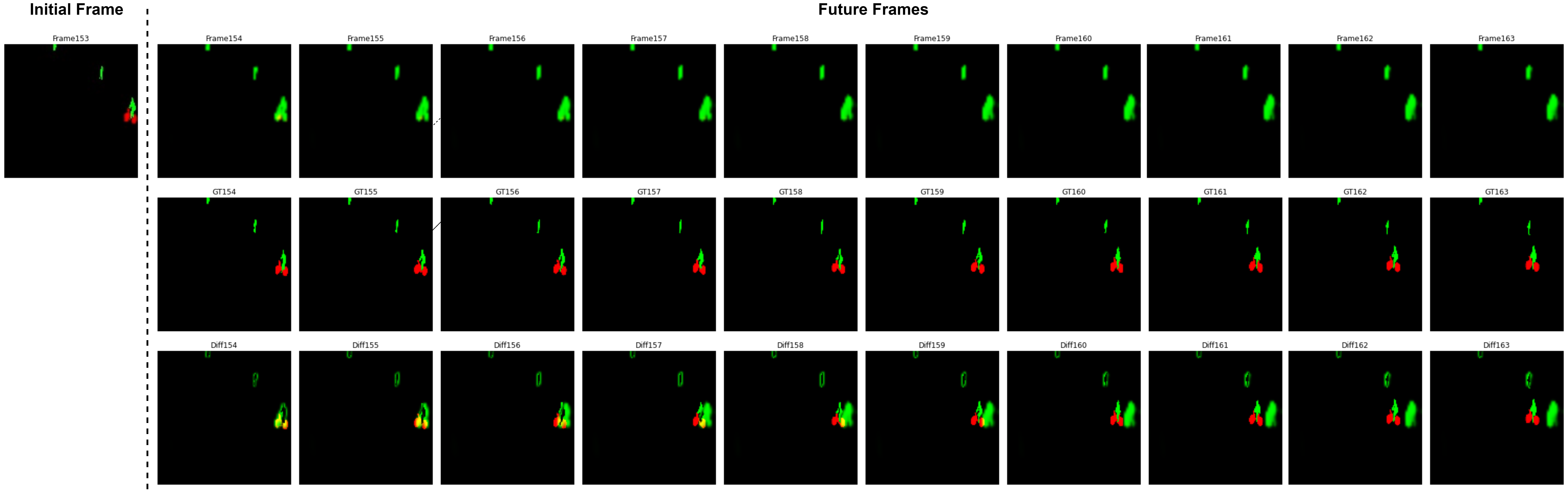}
\caption{Qualitative results for a sample clip from the ShanghaiTech dataset. The first, second, and third rows show the model's predictions, expectations, and related anomaly maps (difference between predictions and expectations), respectively. The first column shows the initial semantic frame and the next frame shows future predictions.} \label{fig5}
\vspace{-3mm}
\end{figure}

Qualitative results show that the proposed Frame-to-Video based method is successful in learning long motion patterns to predict future frame sequences, without generalizing to anomalies (unlike Video-to-Video based methods, which may use motion patterns in the past video to predict future motions). Figure 6 shows similar qualitative results respectively for samples from the UCSD-Ped2 and Ped1.

\begin{figure}[hbt!]
\centering
\includegraphics[width=8.5cm]{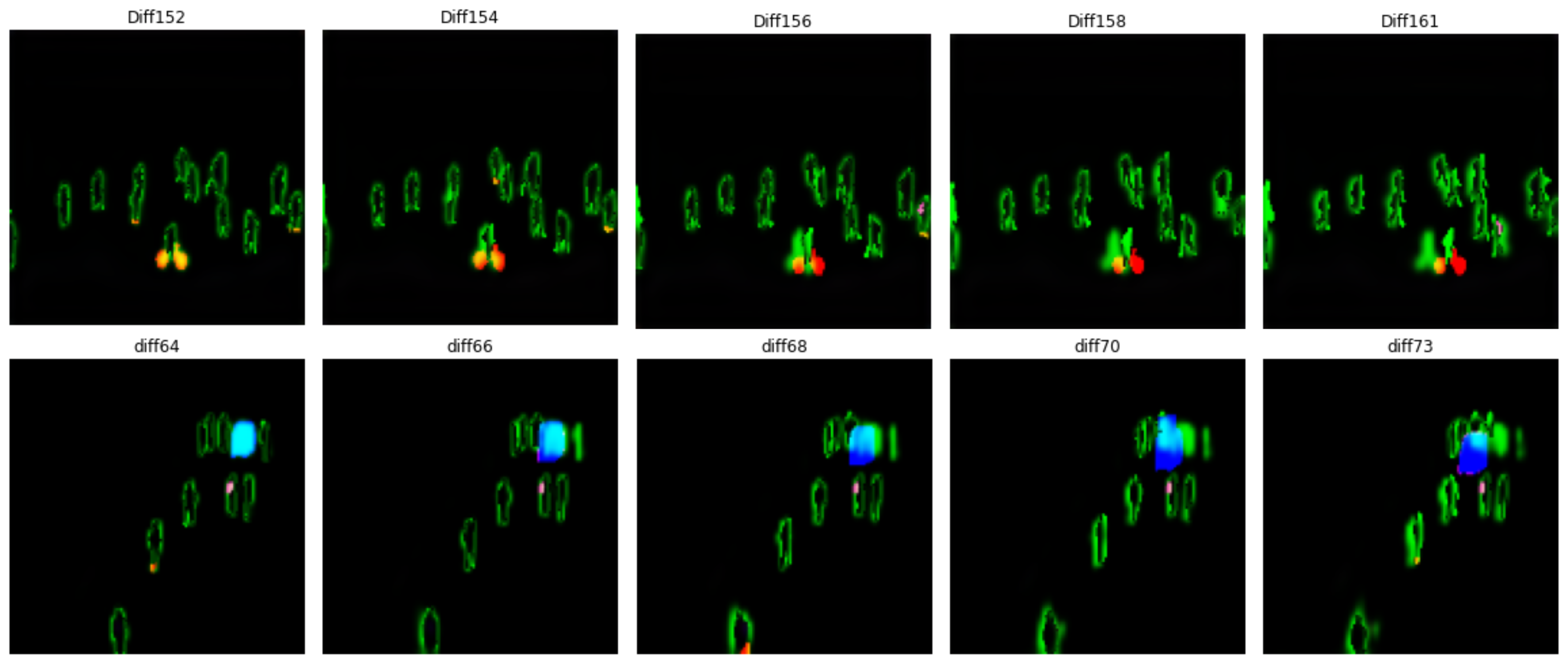}
\caption{Anomaly maps of sample clips from the UCSD-PEd2 and Ped1 respectively in the first and the second rows. Each column shows the anomaly maps of predictions for different frames in the future (columns 1,2,3,4,5 respectively are for the next 1st, 3rd, 5th, 7th, and 10th frames in the future).
} \label{fig6}
\vspace{-3mm}
\end{figure}

%By examining the multiple qualitative results we noticed that, for most of the cases where a human ride a bike or bicycle, the model's future motion estimation for the person might exhibit a partial transformation towards walking. This can be attributed to the fact that the model learns motion patterns while taking into account the specific class of the object. As for this case, the model has predominantly observed walking motion patterns for humans during the training process and considers this pattern for its predictions.
\vspace{-3mm}
%\subsection{Normal motion distribution}
%By examining the KL-divergence for video clips, it can be noticed that the variational encoder has effectively captured the distribution of normal motions. Figure 7, shows the KL-divergence for a sample video clip from the ShanghaiTech dataset. As can be noticed in this figure, the loss value sores as it sees abnormal motion in initial motion maps, as these patterns are out of the normal distribution the model has learned during training.

\subsection{Quantitative evaluation for Frame-to-Video method}
\vspace{-1mm}
Table 2 compares the performance of our proposed method, in terms of AUC, with SOTA prediction-based video anomaly detection methods. This table shows the performances for three benchmark datasets (ShanghaiTech, UCSD-Ped1, and UCSD-Ped2 datasets). As shown in the table, our proposed method shows superior performance for ShanghaiTech and UCSD-Ped2 datasets compared to other holistic prediction-based VAD methods.

\begin{table}[h!]
\begin{center}
\caption{Comparison of performance (AUC) with SOTA work.}\label{tab2}
\scalebox{0.95}{
\begin{tabular}{l|l|l|l}
\hline
Method & ShanghaiTech & UCSD-Ped1 & UCSD-Ped2 \\

\hline\hline
Dong \etal \cite{b14} & 73.7 & N/A   & 95.6 \\
Liu \etal \cite{b31} & 72.8 & 83.1   & 95.4 \\
%Liu \etal \cite{b33} & 76.2 & 91.1   & 99.3 \\
Lu \etal \cite{b35} & N/A & \textbf{86.2}   & 96.06\\
Luo \etal \cite{b38} & 73.0 & 84.3  & 96.2\\
%Ryan \etal \cite{b41} & N/A & N/A & N/A\\
Morais \etal \cite{b43} & 75.4 & N/A  & N/A\\
%Shen \etal \cite{b53} & 79.75 & N/A   & 98.3\\
Ye \etal \cite{b61} & 73.6 & N/A   & 96.8\\
%Zhang \etal \cite{b64} & N/A & N/A   & 95.4\\
%Baradaran \etal \cite{memo2} & 80.61 & N/A & N/A\\
Cai \etal \cite{b4j} & 73.7 & N/A   & 96.6\\
Hui\etal \cite{b27j} & 73.8 & 85.1  & 96.9\\
Park \etal \cite{b47} & 70.5 & N/A   & 97.0\\
Wang \etal \cite{b42j} & 76.6 & 83.4   & 96.3\\
Yang \etal \cite{b47j} & 74.7 & N/A  & 97.6\\
\hline
Ours & \textbf{85.25} & 84.1   & \textbf{97.8}\\
\hline

\end{tabular}
}
\end{center}

\vspace{-4mm}
\end{table}

An examination of Table 2 reveals that the performance of the method is lower for the UCSD-Ped1 dataset. Upon analyzing the anomaly maps for video clips in UCSD-Ped1, we identified a significant number of false positives within the anomaly map. These false positives primarily stem from errors in the performance of Mask-RCNN, due to the low resolution of the UCSD-Ped1 dataset.

\begin{table}
\centering
\caption{Table 3: Performance (AUC) of the Frame-to-Video prediction-based VAD for different prediction time steps in the future. This performance is evaluated on the ShanghaiTech dataset.}\label{tab3}
\begin{tabular}{|l|l|l|l|l|l|l|l|l|l|l|l|}
\hline
Prediction time step &  1 & 2 & 3 & 4 & 5 & 6 & 7 & 8 & 9 & 10 & All\\
\hline
AUC & 83.49 & 84.03 & 85.18 & 86.00 & 86.37 & 86.59 & 86.80 & 86.12 & 85.43 & 84.51 & 85.25\\

\hline
\end{tabular}
\vspace{-2mm}
\end{table}
\vspace{-4mm}
As Table 3 exhibits, the performance of the method (AUC) increases as going toward the future. Although higher performance is achieved when using the prediction error for the next 7th frame in the future, here for the ShanghaiTech dataset, the optimal time step is not always the same across all clips or datasets. On the other hand, the total reconstruction error (mean of the loss for all frames) is always stable in terms of being considerably higher than the loss value of the immediate next frame. Hence, we chose to present this number in our comparisons. This loss also considers all motion patterns (short and long-range patterns) for anomaly detection. Considerable performance rise after the first two frames, as supported by quantitative results, is due to the fact that future frame prediction methods detect motion anomalies in farther frames more effectively, compared to the immediate future frames.

% In addition, our experiments show that the reason for the slight decrease in AUC in higher time steps is the overlapping between estimations of slow motions and expectations of the fast motions.

\vspace{-3mm}
%\subsection{Ablation study}
%\vspace{-1mm}
%In an ablation study, our experiments revealed that by omitting the variational encoder (and hence KL-divergence), predictions get blurry (making detections unexplainable), as expected from deterministic models. Hence, we didn’t calculate AUC for the deterministic model. It is worth mentioning that analyzing the KL loss shows that the variational encoder effectively learns the distribution of normal motions since the related loss value rises in case of motion anomalies (Fig.7). In addition, as shown in Table 4, incorporating time gradient loss in the training brings more attention to motion patterns and thus results in higher performance.

%\vspace{-3mm}
%\begin{figure}[hbt!]
%\centering
%\includegraphics[width=8cm]{kl3.png}
%\caption{KL-divergence for a sample video clip from ShanghaiTech dataset (highlighted frame ranges in red, contain abnormal motion.).} \label{fig7}
%\vspace{-4mm}
%\end{figure}
%\vspace{-3mm}

%\begin{table}
%\centering
%\caption{Contribution of each loss function in the performance of the method.}\label{tab4}
%\begin{tabular}{|l|l|l|l|}
%\hline
%Reconstruction loss &  Time-Gradient loss& KL-divergence & AUC \\
%\hline
%YES & NO & YES & 83.7 \\
%\hline
%YES & YES & YES & 85.25 \\
%\hline
%\end{tabular}
%\vspace{-3mm}
%\end{table}

\section{Conclusion}
\vspace{-2mm}
In this study, we've put forth a new way to detect abnormal patterns in video content, which is based on predicting future video from just one starting frame. To simplify the learning of motion and focus more on the class of objects, we have replaced the original video frames with semantic segmentation maps, which give us a clearer understanding of the objects in the scene. Our experiments have led to two important observations: 1) When we model motion by predicting future video from one frame, we are able to account for longer motions and it is less prone to generalization to anomalies. 2) Predicting video frames further in the future leads to better performance, as shown by a higher AUC. The qualitative results validate our findings, while the quantitative data illustrates the superiority of our proposed method over existing prediction-based VAD methods, particularly when applied to the Shanghaitech and UCSD-Ped2 datasets.

\vspace{-5mm}


\end{document}